\documentclass[a4paper,conference]{IEEEtran}
\IEEEoverridecommandlockouts
\usepackage{hhline}
\usepackage{booktabs}
\usepackage{subfigure}
\usepackage{graphicx}
\usepackage[noend]{algpseudocode}
\usepackage{amsmath} 
\usepackage{algorithmicx,algorithm}
  \graphicspath{{./pdf/}{./}}
  \DeclareGraphicsExtensions{.pdf,.jpeg,.png}

\begin{document}
%
\title{Fused Text Segmentation Networks for Multi-oriented Scene Text Detection}


\author{\IEEEauthorblockN{Yuchen Dai\IEEEauthorrefmark{1},
 Zheng Huang\IEEEauthorrefmark{1}\IEEEauthorrefmark{2},
Yuting Gao\IEEEauthorrefmark{1}, 
Youxuan Xu\IEEEauthorrefmark{3} ,
Kai Chen\IEEEauthorrefmark{1},
Jie Guo\IEEEauthorrefmark{1} and
Weidong Qiu\IEEEauthorrefmark{1}
}
\IEEEauthorblockA{\IEEEauthorrefmark{1}School of Electronic Information and Electrical\\
Shanghai Jiao Tong University,
Shanghai, China}
\IEEEauthorblockA{\IEEEauthorrefmark{2}Westone Cryptologic Research Center,
Beijing, China}
\IEEEauthorblockA{\IEEEauthorrefmark{3}Xiamen No.1 High School,
Fujian, China}

}


%


\maketitle

\begin{abstract}
In this paper, we introduce a novel end-end framework for multi-oriented scene text detection from an instance-aware semantic segmentation perspective. We present Fused Text Segmentation Networks, which combine multi-level features during the feature extracting as text instance may rely on finer feature expression compared to general objects. It detects and segments
the text instance jointly and simultaneously, leveraging merits from both semantic segmentation task and region proposal based object detection task. Not involving any extra pipelines, our approach surpasses the current state of the art on 
multi-oriented scene text detection benchmarks: ICDAR2015 Incidental Scene Text and MSRA-TD500 reaching Hmean 84.1\%  and 82.0\% respectively. Morever, we report a baseline on total-text containing curved text which suggests effectiveness of the proposed approach. 
\end{abstract}


%
\IEEEpeerreviewmaketitle

\section{Introduction}
Recently, scene text detection has drawn great attention from computer vision and machine learning community. Driven by many content-based image applications such as photo translation and receipt content recognition, it has become a promising and challenging research area both in academia and industry. Detecting text in natural images is difficult, because both text and background may be complex in the wild and it often suffers from disturbance such as occlusion and uncontrollable lighting conditions\cite{Zhu2016Scene}.

Previous text detection methods\cite{Chen2004Detecting, epshtein2010detecting,Buta2015FASText,Tian2016Text,Jaderberg2016Reading} have achieved promising results on several benchmarks. The essential problem in text detection is to represent text region using discriminative features. Conventionally, hand-crafted features are designed\cite{epshtein2010detecting,Neumann2012Real,Zamberletti2014Text} to capture the properties of text region such as texture and shape, while in the past few years, deep learning based approaches\cite{Huang2014Robust,Jaderberg2014Deep,Jaderberg2016Reading,Gupta2016Synthetic,Zhang2016Multi,Liao2016TextBoxes} directly learn hierarchical features from training data, demonstrating more accurate and efficient performance in various benchmarks such as ICDAR series contests\cite{Shahab2011ICDAR,Karatzas2013ICDAR,Karatzas2015ICDAR}.

Existing methods\cite{Jaderberg2014Deep,Huang2014Robust,Jaderberg2016Reading,Liao2016TextBoxes} have obtained decent performance for detecting horizontal or near-horizontal text. While horizontal text detection has constraints of axis-aligned bounding-box ground truth, the multi-oriented text is not restrictive to a particular orientation and usually uses quadrilaterals for annotations. Therefore, it reports relatively lower accuracies in ICDAR
2015 Competition Challenge 4 “Incidental scene text localization”\cite{Karatzas2015ICDAR} compared to horizontal scene text detection benchmarks\cite{Shahab2011ICDAR,Karatzas2013ICDAR}.
\begin{figure}[h]
\begin{center}
  
\includegraphics[width=3.5in]{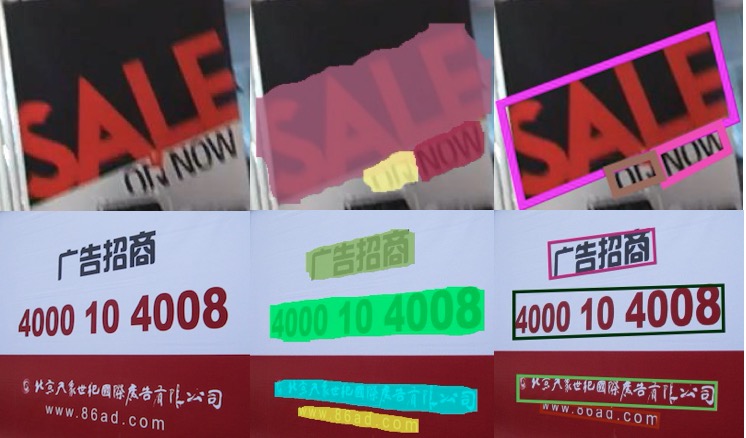}
\caption{\textbf{FTSN Work Flows}. From left to right, input images, text instance segmentation results and final processed quadrilateral results are shown in figure. } 

\end{center}
\end{figure}

Recently, a few approaches\cite{Qin2017Cascaded,He2017deep_direct,Shi2017Detecting,Zhou2017EAST,Liu2017deep,Ma2017arbitrary} have been proposed to address the multi-oriented text
\begin{figure*}[h]
\begin{center}
  
\includegraphics[width=6.2in]{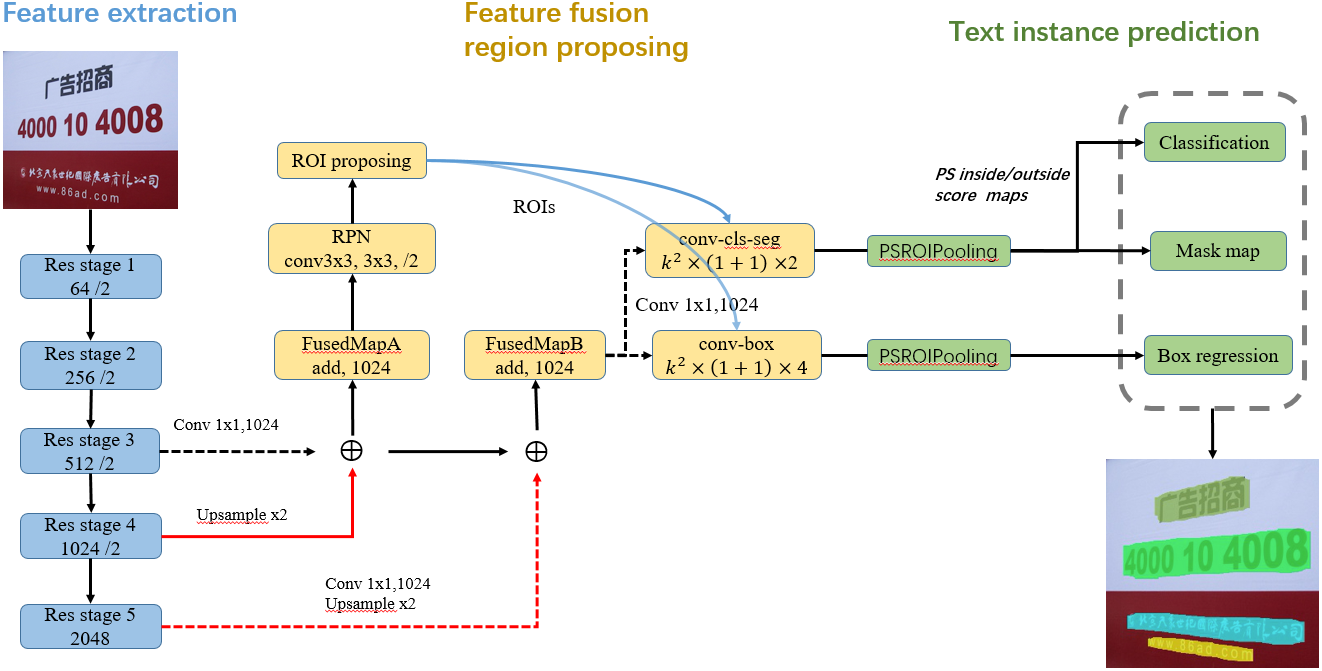}
\caption{
\textbf{The proposed framework} consists of three parts: feature extraction, feature fusion along with region proposing and text instance prediction. The dashed line represents a convolution with 1x1 kernel size and 1024 output channels. The line in red is for upsampling operation and blue lines indicate on which feature maps PSROIPooling are performed using given ROIs.} 

\end{center}
\end{figure*}
detection. In general, there are currently four different types of methods. Region based methods\cite{Shi2017Detecting,Ma2017arbitrary,Liu2017deep} leverage advanced object detection techniques such as Faster RCNN\cite{Ren2017Faster} and SSD\cite{liu2016ssd}. Segmentation-based methods\cite{zhang2016multi_oriented,he2016accurate} mainly utilize fully convolutional neural networks (FCN) for generating text score maps, which often need several stages and components to achieve final detections. Direct regression based method\cite{He2017deep_direct} regresses the position and size of an object from a given point. Finally, hybrid method\cite{Zhou2017EAST} combines text scores map and rotated/quadrangle bounding boxes generation to collaboratively obtain the efficient and accurate performance in multi-oriented text detection. 

Inspired by recent advance of instance-aware semantic  segmentation\cite{li2016fully,he2017mask}, we present a novel perspective to handle the task of multi-oriented text detection. In this work, we leverage the merits from accurate region proposal based methods\cite{Ren2017Faster}, and flexible segmentation based methods which can easily generate arbitrary-shaped text mask\cite{zhang2016multi_oriented,he2016accurate} . It is an end-to-end trainable framework excluding redundant and low-efficient pipelines such as the use of text/nontext
salient map\cite{zhang2016multi_oriented} and text-line generation\cite{he2016accurate}. Based on region proposal network (RPN), our approach detects and segments text instance simultaneously, followed by non-maximum suppression (NMS) to suppress overlapping instances.
Finally, a minimum quadrangle bounding box to fit each instance area is generated as the result of the whole detection process.

Our main contributions are summarized as follows:
\begin{itemize}
\item We present an end-end efficient and  trainable solution for multi-oriented text detection from an instance aware segmentation perspective, excluding any redundant pipelines.
\item During feature extraction, feature maps are composed in a fused fashion to adaptively satisfy the finer representation of text instance.
\item Mask-NMS is introduced to improve the standard NMS when facing heavily inclined or line-level text instances.
\item Without many bells and whistles, our approach outperforms  state of the art on current multi-oriented text detection benchmarks.
\end{itemize}


\section{Related work}
Detecting text in natural images has been widely studied in past few years, motivated by many text-related real-world applications such as photo OCR and blind navigation. One of the mainstream traditional methods for scene text detection are Connected Components (CCs) based methods\cite{neumann2016real,wang2010word,Jaderberg2014Deep,huang2013text,neumann2010method}
 , which consider text as a group of individual components such as  characters. Within these methods, stroke width transform (SWT)\cite{epshtein2010detecting,huang2013text} and maximally stable extremal region (MSER)\cite{Matas2004Robust,neumann2010method,Neumann2012Real} are usually used to seek character candidates. Finally, these candidates are combined to obtain text objects. Although these bottom up approaches may be accurate on some benchmarks\cite{Shahab2011ICDAR,Karatzas2013ICDAR}, they often suffer from too many pipelines, which may cause inefficiency.  
Another mainstream traditional methods are sliding window based\cite{Chen2004Detecting,hanif2009text,Jaderberg2014Deep}. These methods often use a fixed-size or multi-scale window to slide through the image searching the region which most likely contains text.  However, the process of sliding window may involve large computational cost which results in inefficiency. Generally, traditional methods often require several steps to obtain final detections, and hand-designed features are usually used to represent properties of text. 
Therefore, they may suffer from inefficiency and low generalization ability against complex situations such as non-uniform illumination\cite{zhang2016character}.

Recent progress on deep learning based approaches for object detection and semantic segmentation  has provided new techniques for reading text in the wild, which can be also seen as an instance of general object detection. Driven by the advance of object detection frameworks such as Faster RCNN\cite{Ren2017Faster} and SSD\cite{liu2016ssd}, these methods achieved state of the art by either using a region proposal network to first classify some text region proposals\cite{Ma2017arbitrary,Qin2017Cascaded}, or directly regress text
bounding boxes coordinates from a set of default boxes\cite{Liao2016TextBoxes,Shi2017Detecting}.
These methods are able to achieve leading performance on horizontal or multi-oriented scene text detection benchmarks. However, they may also be restricted to rectangular bounding box constraints even with appropriate rotation\cite{Liu2017deep}. Different from these methods, FCN based approaches generate text/non-text map which classifies text at the pixel level\cite{zhang2016multi_oriented}. Though it may be suited well for arbitrary shape of text in natural images, it often involves several pipelines which leads to inefficiency\cite{zhang2016multi_oriented,Qin2017Cascaded}. 

Inspired by recent advance on instance-aware semantic segmentation\cite{li2016fully,he2017mask}, we present an end-end trainable framework called Fused Text Segmentation Networks (FTSN) to handle arbitrary-shape text detection with no extra pipelines involved. It inherits merits from both object detection and semantic segmentation architecture which efficiently detects and segments an text instance simultaneously and accurately gives predictions in the pixel level. As text may rely on finer feature representation, a fused structure formed by multi-level feature maps is set to fit this property.   

\section{Methods}
The proposed framework for multi-oriented scene text detection is diagrammed in Fig.2. It is a deep CNN model which mainly consists of three parts. Feature representations of each image are extracted through resnet-101 backbone\cite{he2016deep}, then multi-level feature maps are fused as FusedMapA which is fed to the region proposed network (RPN) for text region of interest (ROI) generation and FusedMapB for later rois' PSROIPooling. Finally the rois are sent to the detection, segmentation and box regression  branches to output text instances in pixel level along with their corresponding bounding boxes. The post-processing part includes NMS and minimal quadrilateral generation.
\subsection{Network Architecture}
The convolutional feature representation is designed in a fusion fashion. The text instance is not like the general object such as people and cars which have relatively strong semantics. On the  contrary, texts often vary tremendously in intra-class geometries. Consequently, low-level features should be taken into consideration. Basically, resnet-101 consists of five stages. Before region proposing, stage3 and upsampled stage4 feature maps are combined to form FusedMapA through element-wise adding, then upsampled feature maps from stage5 are fused with FusedMapA to form FusedMapB. It is noted that downsampling is not involved during stage5. Instead, we use the “hole algorithm”\cite{chen2016deeplab,long2015fully} to keep the feature stride and maintain the receptive field. The reason for this is that both text properties and the segmentation task may require finer features and involving final downsampling may lose some useful information.

Because using feature stride of stage3 may cause  millions of anchors in original RPN\cite{Ren2017Faster} which makes model training hard, so we add a \begin{math}3\times3\end{math} with stride 2 convolution to reduce such huge number of anchors.

Followed FCIS\cite{li2016fully}, we use Joint Mask Prediction and Classification to simultaneously classify and mask the text instance on \begin{math}2\times(1+1)\end{math} inside/outside score maps generated through PSROIPooling on conv-cls-seg feature maps, and box regression branch utilizes  \begin{math}4\times(1+1)\end{math} feature maps from conv-box after PSROIPooling ("\begin{math}1+1\end{math}" means one class is for text and the other for background). We use  \begin{math}k=7\end{math} shown in Fig.2 in our experiments by default. It is noted that after PSROIPooling, the resolution of feature maps becomes  \begin{math}21\times21\end{math}. Therefore, we use global average pooling\cite{lin2013network} for classification (after pixel-wise max) and box regression branches, and pixel-wise softmax on mask branch.  
\subsection{Ground Truth and Loss Function}
The whole multi-task loss \begin{math}\mathcal{L}\end{math} can be interpreted as 
\begin{equation}
    \mathcal{L} =  \mathcal{L}_{rpn} + \mathcal{L}_{ins}
\end{equation}
\begin{equation}
    \mathcal{L}_{rpn} =  \mathcal{L}_{rcls} + \lambda_{r}\mathcal{L}_{rbox}
\end{equation}
\begin{equation}
    \mathcal{L}_{ins} =  \mathcal{L}_{cls} + \lambda_{m}\mathcal{L}_{mask} +  \lambda_{b}\mathcal{L}_{box}
\end{equation}

The full loss \begin{math}\mathcal{L}\end{math} consists of two sub stage losses: RPN loss \begin{math}\mathcal{L}_{rpn}\end{math} where \begin{math}\mathcal{L}_{rcls}\end{math} is for region proposal classification and \begin{math}\mathcal{L}_{rbox}\end{math} is for box regression, and text instance loss \begin{math}\mathcal{L}_{ins}\end{math} based on each ROI, where \begin{math}\mathcal{L}_{cls},\mathcal{L}_{mask} and \mathcal{L}_{box}\end{math} represent losses for instance classification, mask and box regression task respectively. \begin{math}\lambda\end{math} is the hyper-parameter to control the balance among each loss term. They are set as  \begin{math}\lambda_{r}=0.2, \lambda_{m}=2, \lambda_{b}=0.2\end{math} in our experiments.

Classification and mask task both use cross-entropy as loss function, whereas we use smooth-L1 for box regression task formulated as 

\begin{equation}
\text{smooth}_{L_1}\left(x\right) = 
\left\{\begin{matrix}
0.5(\sigma x)^{2} & \text{if} \left | x \right |  < 1/\sigma^{2} , \\ 
\left |x\right | - 0.5/\sigma^{2} & \text{otherwise} .
\end{matrix}\right.
\end{equation}

\begin{math}\sigma\end{math} is set to 3 in our experiments which makes the box regression loss less sensitive to outliers. 

\begin{figure}
  \centering
  \subfigure{
    \label{fig:subfig:a} 
    \includegraphics[width=1.5in]{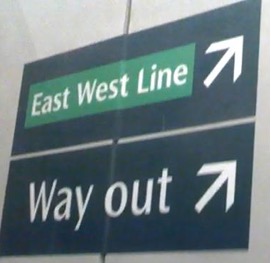}}
  \hspace{0.1in}
  \subfigure{
    \label{fig:subfig:b} 
    \includegraphics[width=1.5in]{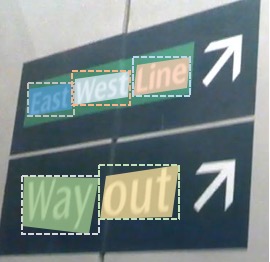}}
  \caption{\textbf{Ground truth.} Left: original image. Right: corresponding ground truth in which dashed lines are for bounding boxes and quadrilaterals filled with different colors are for masks}
  \label{fig:subfig} 
\end{figure} 
Ground truth of each text instance is presented by bounding boxes and masks shown in Fig.3. In most multi-oriented text detection dataset, annotations are given in quadrilaterals such as IC15 or can be converted to quadrilaterals such as TD500. For each instance, we directly generate mask from quadrilateral coordinates and use the minimal rectangle containing the mask as the bounding box.
\begin{figure}[h]
\begin{center}
  
\includegraphics[width=3.5in]{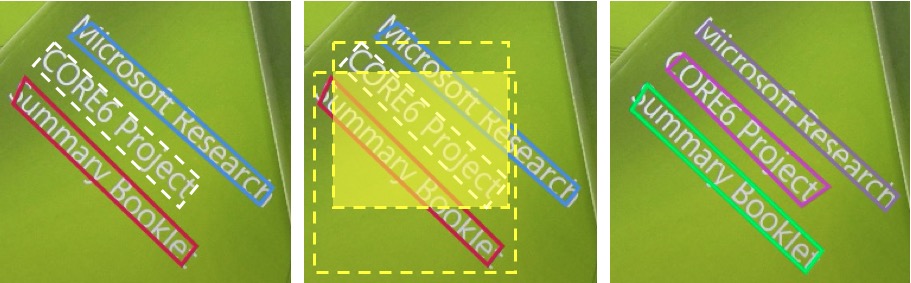}
\caption{\textbf{Heavily inclined text filtering results.} Left: standard NMS may filter the correct detection as shown in white dashed line. Middle: dashed lines in yellow show the bounding boxes used in IOU computation of standard NMS, and the yellow transparent area illustrates that the IOU between the two closed inclined instance is big enough to make the correct detection filtered. Right: Mask-NMS takes the mask area of each instance to compute MMI replacing IOU. Therefore, there is no intersection between each instance. } 

\end{center}
\end{figure}

\subsection{Post Processing}
\textbf{Mask-NMS} To obtain final detection results, we use Non-Maximum Suppression mechanism (NMS) to filter overlapped text instances and preserve those with highest scores. After NMS, we generate a minimum quadrilateral for each text instance covering the mask as shown in Fig.1. 

Standard NMS computes IOU among bounding boxes, which may be fine for word-level and near-horizontal  results' filtering. However, it may filter some correct line-level detections when they are close and heavily inclined as shown in Fig.4 or when words stay close in the same line as shown in Fig.5. Consequently, we propose a modified NMS called Mask-NMS to handle such situations. Mask-NMS mainly changes bounding box IOU computation to so-called mask-maximum-intersection (MMI) as formulated:
\begin{equation}
    MMI =  max(I/I_{A}, I/I_{B})
\end{equation}
\begin{math}I_{A},I_{B}\end{math} are mask areas of two text instances to be computed,  \begin{math}I\end{math} is the intersection area between the masks. Maximum intersection over the mask areas are used to replace original IOU for the reason that detections may easily involve line-level and word-level text instances simultaneously at the same line as shown in Fig.5. 
The proposed Mask-NMS has significantly improved performance for multi-oriented scene text detection as shown in section.5. 
\begin{figure}[h]
\begin{center}
  
\includegraphics[width=3.5in]{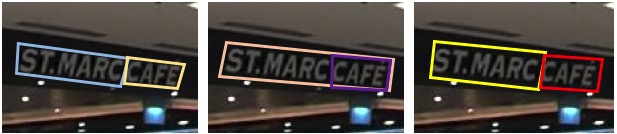}
\caption{\textbf{line-level text NMS results}. From left to right: ground truth, standard NMS  and Mask-NMS results. Benchmarks may provide annotations at different levels such as word-level in IC15 and line-level in TD500. However, model may be confused about them and make predictions at different levels as shown in the middle. Using maximum intersection can greatly avoid this situation as the MMI over the two text instances in the middle image is 1 so one of them is certainly filtered.} 

\end{center}
\end{figure}
\begin{figure*}
\begin{center}
  
\includegraphics[width=7in]{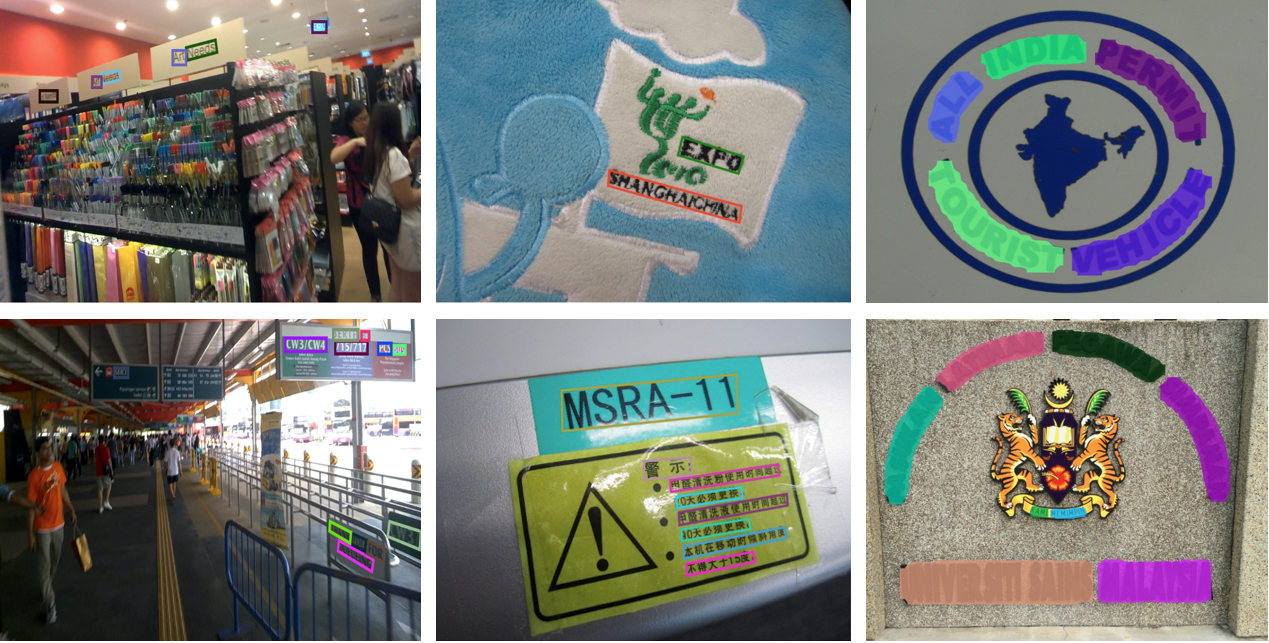}
\caption{Results of ICDAR2015, MSRA-TD500 and Total-Text} 

\end{center}
\end{figure*} 
\section{Experiments}
To evaluate the proposed framework, we conduct quantitative experiments
on three public benchmarks: ICDAR2015, MSRA-TD500 and Total-Text.
\subsection{Datasets}
\textbf{ICDAR 2015 Incidental Text (IC15)} the Challenge 4 of ICDAR 2015 Robust
Reading Competition\cite{Karatzas2015ICDAR}. IC15 contains 1000 training and 500 testing incidental images taken by Google Glasses without paying attention to viewpoint and image quality. Therefore, large variations in text scale, orientation and resolution lead to difficulty for text detection. Annotations of the dataset are given in word-level
quadrilaterals.

\textbf{MSRA-TD500 (TD500)} is early presented in \cite{yao2012detecting}. The dataset is multi-oriented and multi-lingual including both Chinese and English text， which consists of 300 training and 200 testing images. Different from IC15, annotations of TD500 are at line level which are rotated rectangles.

\textbf{Total-Text} is presented in ICDAR2017\cite{CK2017}. It consists of 1555 images with more than 3 different text orientations: Horizontal, Multi-Oriented, and Curved.

\textbf{SynthText in the Wild (SynthText)} The dataset contains 800,000 synthetic images\cite{Gupta2016Synthetic}, text with random color, fonts, scale and orientation are rendered on natural images carefully to have a realistic look. Annotations are given in character, word and line level. 

\subsection{Implementation Details}
\textbf{Training} We pretrain the proposed FTSN on a subset of SynthText containing 160,000 images, then finetune on IC15, TD500 and Total-Text. For optimization, standard SGD is used during training with learning rate \begin{math}5\times10^{-3}\end{math} for first 5 epochs and \begin{math}5\times10^{-4}\end{math} for the last epoch,and we also apply online  hard  example  mining (OHEM)\cite{shrivastava2016training} for balancing the positive and negative samples. Different from original RPN anchor ratios and scales setting for object detection, anchor scales of [\begin{math}32^{2},64^{2},128^{2},256^{2}\end{math}] and ratios of [1/3,1/2,1,2,3,5,7] are set because text often has a large aspect ratio and a small scale\cite{Liao2016TextBoxes}.

\textbf{Data augmentation} Multi-scale training, rotation and color jittering are applied during training. Scales are randomly chosen from [600,720,960,1100] and each number represents the short edge of input images. Rotation with $15^{\circ}$ ,$30^{\circ}$  and $45^{\circ}$  are applied with horizontal flip. Consequently, it enlarges 8x dataset size than the original one. Random brightness, contrast and saturation jittering are applied for input images.

\textbf{Testing} Input images are resized to \begin{math}848\times1500\end{math} when testing. After NMS, mask voting\cite{dai2016instance} is used to obtain an ensemble text instance mask by averaging all reasonable detections. 

Experiments are conducted on MXNet\cite{chen2015mxnet} and run on a server with Intel i7 6700K CPU, 64GB RAM, GTX 1080 and Ubuntu 14.04 OS.
\begin{table}[htbp]
 \caption{\label{tab:1} ICDAR 2015 incidental dataset}
 \begin{center}
\begin{tabular}
{c||c|c|c}
Method & Precision (\%) & Recall (\%) & Hmean (\%) \\
\hline\hline
 HUST\cite{Karatzas2015ICDAR} & 44.0 & 37.8 & 40.7 \\
 Zhang \emph{et al.}\cite{zhang2016multi_oriented} & 71.0 & 43.0 & 54.0\\
 DMPNet\cite{Liu2017deep} & 73.2 & 68.2 & 70.6 \\
Qin \emph{et al.}\cite{Qin2017Cascaded} & 79.0 & 65.0 & 71.0 \\
SegLink\cite{Shi2017Detecting}  & 73.1 & 76.8 & 75.0 \\
RRPN\cite{Ma2017arbitrary} & 73.2 & 82.2 & 77.4 \\
EAST\cite{Zhou2017EAST} & 83.3 & 78.3 & 80.7 \\
He \emph{et al.}\cite{He2017deep_direct} & 82.0 & 80.0 & 81.0\\
\hline\hline
\textbf{Proposed FTSN+SNMS} & 87.1 & 80.0 & 83.4 \\
\textbf{Proposed FTSN+MNMS} & \textbf{88.6} & 80.0 & \textbf{84.1} \\
\hline
 \end{tabular}
 \end{center}
\end{table}
\begin{table}[htbp]
 \caption{\label{tab:2} MSRA-TD500 dataset}
 \begin{center}
\begin{tabular}
{c||c|c|c}
Method & Precision (\%) & Recall (\%) & Hmean (\%) \\
\hline\hline
 Yao \emph{et al.}\cite{yao2012detecting} & 63.0 & 63.0 & 60.0 \\
 Zhang \emph{et al.}\cite{zhang2016multi_oriented} & 83.0 & 67.0 & 74.0\\
RRPN\cite{Ma2017arbitrary} & 82.0 & 68.0 & 74.0\\
He \emph{et al.}\cite{He2017deep_direct} & 77.0 & 70.0 & 74.0\\
EAST\cite{Zhou2017EAST} & 87.3 & 67.4 & 76.1 \\
SegLink\cite{Shi2017Detecting}  & 86.0 & 70.0 & 77.0 \\
\hline\hline
\textbf{Proposed FTSN+SNMS} & 86.6 & \textbf{77.3} & 81.7 \\
\textbf{Proposed FTSN+MNMS} & \textbf{87.6} & 77.1 & \textbf{82.0} \\
\hline
 \end{tabular}
 \end{center}
\end{table}

\begin{table}[htbp]
 \caption{\label{tab:3} Total-Text dataset}
 \begin{center}
\begin{tabular}
{c||c|c|c}
Method & Precision (\%) & Recall (\%) & Hmean (\%) \\
\hline\hline
\textbf{Proposed FTSN} & 84.7 & 78.0 & 81.3 \\
\hline
 \end{tabular}
 \end{center}
\end{table}
\subsection{Results}
Tabel.1 shows results of the proposed FTSN on IC15 compared with previous state of art published methods. SNMS and MNMS represent standard NMS and Mask-NMS respectively. Our FTSN with Mask-NMS outperforms former best result by 5.3\% in Precision and 3.1\% in Hmean. It is evaluated by the official submission server\footnote{http://rrc.cvc.uab.es/?ch=4}. 

Results on TD500 are shown in Table.2 along with other state of art methods. It is shown that our methods outperform the current state of art approaches by a large margin in Hmean and Recall, without adding extra real-world training images. 

Our method also shows great flexibility on the total-text dataset containing curved text. As the dataset is new to the community, experiments are seldom conducted on it which makes our results as a baseline shown in Table3. The evaluation metric uses IoU of 0.5 between each instance masks. 

Outperforming the current state of the art, our approach runs about 4 FPS on \begin{math}848\times1500\end{math} images and 2.5 FPS when using Mask-NMS, which presents efficiency and accuracy.

It is noted that the proposed Mask-NMS significantly improved Hmean by 0.7 and 0.3 percent on IC15 and TD500, which mainly target the situations in Fig.4 and Fig.5. 

Fig.6 shows example results of FTSN. From left to right, it illustrates results on IC15 ,TD500 and Total-Text dataset. The decent performance for word-level, line-level and curved text detection with large variation in resolution, view point, scale and linguistics suggests excellent generalization ability.

\section{Conclusion}
We present FTSN, an end-end efficient and accurate multi-oriented scene text detection framework. It has outperformed previous state of the art approaches on word-level line-level annotated benchmarks and report a baseline on total-text demonstrating decent generalization ability and flexibility. 

\section{Acknowlegements}
The research is supported by The National Key Research and Development Program of China under grant 2017YFB1002401.

\bibliographystyle{./IEEEtran}
\bibliography{main}

\begin{thebibliography}{10}
\providecommand{\url}[1]{#1}
\csname url@samestyle\endcsname
\providecommand{\newblock}{\relax}
\providecommand{\bibinfo}[2]{#2}
\providecommand{\BIBentrySTDinterwordspacing}{\spaceskip=0pt\relax}
\providecommand{\BIBentryALTinterwordstretchfactor}{4}
\providecommand{\BIBentryALTinterwordspacing}{\spaceskip=\fontdimen2\font plus
\BIBentryALTinterwordstretchfactor\fontdimen3\font minus
  \fontdimen4\font\relax}
\providecommand{\BIBforeignlanguage}[2]{{%
\expandafter\ifx\csname l@#1\endcsname\relax
\typeout{** WARNING: IEEEtran.bst: No hyphenation pattern has been}%
\typeout{** loaded for the language `#1'. Using the pattern for}%
\typeout{** the default language instead.}%
\else
\language=\csname l@#1\endcsname
\fi
#2}}
\providecommand{\BIBdecl}{\relax}
\BIBdecl

\bibitem{Zhu2016Scene}
Y.~Zhu, C.~Yao, and X.~Bai, ``Scene text detection and recognition: recent
  advances and future trends,'' \emph{Frontiers of Computer Science}, vol.~10,
  no.~1, pp. 19--36, 2016.

\bibitem{Chen2004Detecting}
X.~Chen and A.~L. Yuille, ``Detecting and reading text in natural scenes,'' in
  \emph{IEEE Computer Society Conference on Computer Vision and Pattern
  Recognition}, 2004, pp. 366--373.

\bibitem{epshtein2010detecting}
B.~Epshtein, E.~Ofek, and Y.~Wexler, ``Detecting text in natural scenes with
  stroke width transform,'' in \emph{Computer Vision and Pattern Recognition
  (CVPR), 2010 IEEE Conference on}.\hskip 1em plus 0.5em minus 0.4em\relax
  IEEE, 2010, pp. 2963--2970.

\bibitem{Buta2015FASText}
M.~Buta, L.~Neumann, and J.~Matas, ``Fastext: Efficient unconstrained scene
  text detector,'' in \emph{IEEE International Conference on Computer Vision},
  2015, pp. 1206--1214.

\bibitem{Tian2016Text}
S.~Tian, Y.~Pan, C.~Huang, S.~Lu, K.~Yu, and C.~L. Tan, ``Text flow: A unified
  text detection system in natural scene images,'' in \emph{IEEE International
  Conference on Computer Vision}, 2016, pp. 4651--4659.

\bibitem{Jaderberg2016Reading}
M.~Jaderberg, K.~Simonyan, A.~Vedaldi, and A.~Zisserman, ``Reading text in the
  wild with convolutional neural networks,'' \emph{International Journal of
  Computer Vision}, vol. 116, no.~1, pp. 1--20, 2016.

\bibitem{Neumann2012Real}
L.~Neumann and J.~Matas, ``Real-time scene text localization and recognition,''
  in \emph{IEEE Conference on Computer Vision and Pattern Recognition}, 2012,
  pp. 3538--3545.

\bibitem{Zamberletti2014Text}
A.~Zamberletti, L.~Noce, and I.~Gallo, \emph{Text Localization Based on Fast
  Feature Pyramids and Multi-Resolution Maximally Stable Extremal
  Regions}.\hskip 1em plus 0.5em minus 0.4em\relax Springer International
  Publishing, 2014.

\bibitem{Huang2014Robust}
W.~Huang, Y.~Qiao, and X.~Tang, \emph{Robust Scene Text Detection with
  Convolution Neural Network Induced MSER Trees}.\hskip 1em plus 0.5em minus
  0.4em\relax Springer International Publishing, 2014.

\bibitem{Jaderberg2014Deep}
M.~Jaderberg, A.~Vedaldi, and A.~Zisserman, ``Deep features for text
  spotting,'' in \emph{European Conference on Computer Vision}, 2014, pp.
  512--528.

\bibitem{Gupta2016Synthetic}
A.~Gupta, A.~Vedaldi, and A.~Zisserman, ``Synthetic data for text localisation
  in natural images,'' pp. 2315--2324, 2016.

\bibitem{Zhang2016Multi}
Z.~Zhang, C.~Zhang, W.~Shen, C.~Yao, W.~Liu, and X.~Bai, ``Multi-oriented text
  detection with fully convolutional networks,'' in \emph{Computer Vision and
  Pattern Recognition}, 2016.

\bibitem{Liao2016TextBoxes}
M.~Liao, B.~Shi, X.~Bai, X.~Wang, and W.~Liu, ``Textboxes: A fast text detector
  with a single deep neural network,'' in \emph{Association for the Advancement
  of Artificial Intelligence}, 2017.

\bibitem{Shahab2011ICDAR}
A.~Shahab, F.~Shafait, and A.~Dengel, ``Icdar 2011 robust reading competition
  challenge 2: Reading text in scene images,'' in \emph{International
  Conference on Document Analysis and Recognition}, 2011, pp. 1491--1496.

\bibitem{Karatzas2013ICDAR}
D.~Karatzas, F.~Shafait \emph{et~al.}, ``Icdar 2013 robust reading
  competition,'' in \emph{International Conference on Document Analysis and
  Recognition}, 2013, pp. 1484--1493.

\bibitem{Karatzas2015ICDAR}
D.~Karatzas, L.~Gomez-Bigorda \emph{et~al.}, ``Icdar 2015 competition on robust
  reading,'' in \emph{International Conference on Document Analysis and
  Recognition}, 2015, pp. 1156--1160.

\bibitem{Qin2017Cascaded}
S.~Qin and R.~Manduchi, ``Cascaded segmentation-detection networks for
  word-level text spotting,'' 2017.

\bibitem{He2017deep_direct}
W.~He, X.-Y. Zhang, F.~Yin, and C.-L. Liu, ``Deep direct regression for
  multi-oriented scene text detection,'' \emph{arXiv preprint
  arXiv:1703.08289}, 2017.

\bibitem{Shi2017Detecting}
B.~Shi, X.~Bai, and S.~Belongie, ``Detecting oriented text in natural images by
  linking segments,'' in \emph{Computer Vision and Pattern Recognition}, 2017.

\bibitem{Zhou2017EAST}
X.~Zhou, C.~Yao, H.~Wen, Y.~Wang, S.~Zhou, W.~He, and J.~Liang, ``East: An
  efficient and accurate scene text detector,'' in \emph{Computer Vision and
  Pattern Recognition}, 2017.

\bibitem{Liu2017deep}
Y.~Liu and L.~Jin, ``Deep matching prior network: Toward tighter multi-oriented
  text detection,'' \emph{arXiv preprint arXiv:1703.01425}, 2017.

\bibitem{Ma2017arbitrary}
J.~Ma, W.~Shao, H.~Ye, L.~Wang, H.~Wang, Y.~Zheng, and X.~Xue,
  ``Arbitrary-oriented scene text detection via rotation proposals,''
  \emph{arXiv preprint arXiv:1703.01086}, 2017.

\bibitem{Ren2017Faster}
S.~Ren, K.~He, R.~Girshick, and J.~Sun, ``Faster r-cnn: Towards real-time
  object detection with region proposal networks.'' \emph{IEEE Transactions on
  Pattern Analysis \& Machine Intelligence}, vol.~39, no.~6, p. 1137, 2017.

\bibitem{liu2016ssd}
W.~Liu, D.~Anguelov, D.~Erhan, C.~Szegedy, S.~Reed, C.-Y. Fu, and A.~C. Berg,
  ``Ssd: Single shot multibox detector,'' in \emph{European conference on
  computer vision}.\hskip 1em plus 0.5em minus 0.4em\relax Springer, 2016, pp.
  21--37.

\bibitem{zhang2016multi_oriented}
Z.~Zhang, C.~Zhang, W.~Shen, C.~Yao, W.~Liu, and X.~Bai, ``Multi-oriented text
  detection with fully convolutional networks,'' in \emph{Proceedings of the
  IEEE Conference on Computer Vision and Pattern Recognition}, 2016, pp.
  4159--4167.

\bibitem{he2016accurate}
T.~He, W.~Huang, Y.~Qiao, and J.~Yao, ``Accurate text localization in natural
  image with cascaded convolutional text network,'' \emph{arXiv preprint
  arXiv:1603.09423}, 2016.

\bibitem{li2016fully}
Y.~Li, H.~Qi, J.~Dai, X.~Ji, and Y.~Wei, ``Fully convolutional instance-aware
  semantic segmentation,'' in \emph{Proceedings of the IEEE Conference on
  Computer Vision and Pattern Recognition}, 2017.

\bibitem{he2017mask}
K.~He, G.~Gkioxari, P.~Doll{\'a}r, and R.~Girshick, ``Mask r-cnn,'' \emph{arXiv
  preprint arXiv:1703.06870}, 2017.

\bibitem{neumann2016real}
L.~Neumann and J.~Matas, ``Real-time lexicon-free scene text localization and
  recognition,'' \emph{IEEE transactions on pattern analysis and machine
  intelligence}, vol.~38, no.~9, pp. 1872--1885, 2016.

\bibitem{wang2010word}
K.~Wang and S.~Belongie, ``Word spotting in the wild,'' in \emph{European
  Conference on Computer Vision}.\hskip 1em plus 0.5em minus 0.4em\relax
  Springer, 2010, pp. 591--604.

\bibitem{huang2013text}
W.~Huang, Z.~Lin, J.~Yang, and J.~Wang, ``Text localization in natural images
  using stroke feature transform and text covariance descriptors,'' in
  \emph{Proceedings of the IEEE International Conference on Computer Vision},
  2013, pp. 1241--1248.

\bibitem{neumann2010method}
L.~Neumann and J.~Matas, ``A method for text localization and recognition in
  real-world images,'' in \emph{Asian Conference on Computer Vision}.\hskip 1em
  plus 0.5em minus 0.4em\relax Springer, 2010, pp. 770--783.

\bibitem{Matas2004Robust}
J.~Matas, O.~Chum, M.~Urban, and T.~Pajdla, ``Robust wide-baseline stereo from
  maximally stable extremal regions,'' \emph{Image \& Vision Computing},
  vol.~22, no.~10, pp. 761--767, 2004.

\bibitem{hanif2009text}
S.~M. Hanif and L.~Prevost, ``Text detection and localization in complex scene
  images using constrained adaboost algorithm,'' in \emph{Document Analysis and
  Recognition, 2009. ICDAR'09. 10th International Conference on}.\hskip 1em
  plus 0.5em minus 0.4em\relax IEEE, 2009, pp. 1--5.

\bibitem{zhang2016character}
S.~Zhang, M.~Lin, T.~Chen, L.~Jin, and L.~Lin, ``Character proposal network for
  robust text extraction,'' in \emph{Acoustics, Speech and Signal Processing
  (ICASSP), 2016 IEEE International Conference on}.\hskip 1em plus 0.5em minus
  0.4em\relax IEEE, 2016, pp. 2633--2637.

\bibitem{he2016deep}
K.~He, X.~Zhang, S.~Ren, and J.~Sun, ``Deep residual learning for image
  recognition,'' in \emph{Proceedings of the IEEE conference on computer vision
  and pattern recognition}, 2016, pp. 770--778.

\bibitem{chen2016deeplab}
L.-C. Chen, G.~Papandreou, I.~Kokkinos, K.~Murphy, and A.~L. Yuille, ``Deeplab:
  Semantic image segmentation with deep convolutional nets, atrous convolution,
  and fully connected crfs,'' \emph{arXiv preprint arXiv:1606.00915}, 2016.

\bibitem{long2015fully}
J.~Long, E.~Shelhamer, and T.~Darrell, ``Fully convolutional networks for
  semantic segmentation,'' in \emph{Proceedings of the IEEE Conference on
  Computer Vision and Pattern Recognition}, 2015, pp. 3431--3440.

\bibitem{lin2013network}
M.~Lin, Q.~Chen, and S.~Yan, ``Network in network,'' \emph{arXiv preprint
  arXiv:1312.4400}, 2013.

\bibitem{yao2012detecting}
C.~Yao, X.~Bai, W.~Liu, Y.~Ma, and Z.~Tu, ``Detecting texts of arbitrary
  orientations in natural images,'' in \emph{Computer Vision and Pattern
  Recognition (CVPR), 2012 IEEE Conference on}.\hskip 1em plus 0.5em minus
  0.4em\relax IEEE, 2012, pp. 1083--1090.

\bibitem{CK2017}
C.~K. Ch’ng and C.~S. Chan, ``Total-text: A comprehensive dataset for scene
  text detection and recognition,'' in \emph{14th IAPR International Conference
  on Document Analysis and Recognition {ICDAR}}, 2017.

\bibitem{shrivastava2016training}
A.~Shrivastava, A.~Gupta, and R.~Girshick, ``Training region-based object
  detectors with online hard example mining,'' in \emph{Proceedings of the IEEE
  Conference on Computer Vision and Pattern Recognition}, 2016, pp. 761--769.

\bibitem{dai2016instance}
J.~Dai, K.~He, and J.~Sun, ``Instance-aware semantic segmentation via
  multi-task network cascades,'' in \emph{Proceedings of the IEEE Conference on
  Computer Vision and Pattern Recognition}, 2016, pp. 3150--3158.

\bibitem{chen2015mxnet}
T.~Chen, M.~Li, Y.~Li, M.~Lin, N.~Wang, M.~Wang, T.~Xiao, B.~Xu, C.~Zhang, and
  Z.~Zhang, ``Mxnet: A flexible and efficient machine learning library for
  heterogeneous distributed systems,'' \emph{arXiv preprint arXiv:1512.01274},
  2015.

\end{thebibliography}
\end{document}